# Novel Dynamic Batch-Sensitive Adam Optimiser for Vehicular Accident Injury Severity Prediction.

**Daniel Asare Kyei, Alimatu Saadia-Yussiff, Maame G. Asante-Mensah *, Abdul Lateef-Yussiff, Charles Roland Haruna, Derry Emmanuel.**

Department of Computer Science and Information Technology, University of Cape Coast, Cape Coast, Ghana;
Email: gasante-mensah@ucc.edu.gh

## ABSTRACT

The choice of optimiser is important in deep learning, as it strongly influences model efficiency and speed of convergence. However, many commonly used optimisers encounter difficulties when applied to imbalanced and sequential datasets, limiting their ability to capture patterns of minority classes. In this study, we propose Dynamic Batch-Sensitive Adam (DBS-Adam), an optimiser that dynamically scales the learning rate using a batch difficulty score derived from exponential moving averages of gradient norms and batch loss. DBS-Adam improves training stability and accelerates convergence by increasing updates for difficult batches and reducing them for easier ones. We evaluate DBS-Adam by integrating it with Bi-Directional LSTM networks for accident injury severity prediction, addressing class imbalance through SMOTE-ENN resampling and Focal Loss. Four experimental configurations compare baseline Bi-LSTM models and alternative architectures to assess optimiser impact. Rigorous comparison against state-of-the-art optimisers (AMSGrad, AdamW, AdaBound) across five random seeds demonstrated DBS-Adam's competitive performance with statistically significant precision improvements ($p = 0.020$). Results indicate that DBS-Adam outperforms standard optimisation approaches, achieving 95.22% test accuracy, 96.11% precision, 95.28% recall, 95.39% F1-score, and a test loss of 0.0086. The proposed framework enables effective real-time accident severity classification for targeted emergency response and road safety interventions, demonstrating the value of DBS-Adam for learning from imbalanced sequential data.

## 1. Introduction

The performance of deep learning models is heavily influenced by the optimisation technique employed in training. Stochastic gradient descent (SGD), Adam, and RMSProp are examples of common optimisers that may experience slow convergence, instability, and decreased efficacy when used on noisy, unbalanced, or sequential datasets [1], [2], [3], [4]. Applications that depend on safety are most affected by these flaws. Because they combine temporal structure, measurement noise, and a noticeable class imbalance, road accident records are a prime example of these difficulties, making it more challenging to train trustworthy models.

Road traffic accidents (RTAs) represent a critical global health burden: the World Health Organisation estimates about 1.3 million deaths in 2021 ($\approx$15 deaths per 100,000 population), with the greatest impact felt in low- and middle-income countries. Accurate prediction of injury severity from RTAs using machine learning (ML) and deep learning (DL) methods can improve emergency response, reduce mortality, and enable more efficient allocation of medical resources. Such predictive systems can inform triage decisions, prioritise ambulance dispatch, and guide short-term resource planning at crash sites and receiving hospitals [5], [6], [7]. Recent work has shown the usefulness of neural networks for accident prediction. [8] reported that an LSTM-based RNN achieved 71.77% accuracy, outperforming a multilayer perceptron (65.48%) and Bayesian logistic regression (58.30%) on Malaysian accident data. [9] applied an LSTM with SMOTE for real-time crash risk prediction, reaching a sensitivity of 60.67% compared with 56.72% for a conditional logistic model. These studies confirm the potential of ANN and LSTM methods for accident analysis but also highlight continuing difficulties with class imbalance and reliance on conventional optimisers, pointing to the need for improved optimisation strategies such as the DBS-Adam proposed in this study.

To address some of these optimisation challenges in imbalanced sequential domains, this study introduces Dynamic Batch-Sensitive Adam (DBS-Adam), a novel optimiser that adapts the global learning rate at the batch level using a difficulty score derived from exponential moving averages of gradient norms and batch loss. DBS-Adam is designed to stabilise training and accelerate convergence for deep models trained on imbalanced traffic-accident data. We evaluate the optimiser by integrating it with a Bi-Directional Long Short-Term Memory (Bi-LSTM) network for injury-severity prediction and demonstrate its effectiveness in overcoming key optimisation issues in this domain.

## 1.1. Traditional Statistical Methods vs. Machine Learning Approaches

Traditional statistical approaches, such as logistic regression and generalised linear models, have long been used to analyse crash injury severity. Studies employing logistic regression report accuracies of up to 87% [10], while generalised linear models achieve coefficients of determination of approximately 0.62 [11]. Time-series techniques such as ARIMA have also been applied, outperforming Poisson and negative binomial models in predicting accident frequency in Malaysian datasets [12]. These approaches, however, rely on strong distributional assumptions and often struggle to capture complex interactions in heterogeneous crash environments.

In contrast, machine learning methods have demonstrated improved predictive capability, particularly when modelling non-linear relationships and high-dimensional feature spaces. Techniques such as Random Forests, XGBoost, Support Vector Machines, multilayer perceptrons and naïve Bayes classifiers have shown substantial gains in accuracy and sensitivity across various accident-severity studies [13], [14], [15], [16], [17]. Neural network-based approaches, including LSTM and Bi-LSTM architectures, have further improved performance in sequential crash data by modelling temporal dependencies and contextual patterns. The increasing adoption of these methods reflects their flexibility and stronger empirical performance compared with traditional statistical models.

## 1.2. Artificial Neural Network Architectures

The most commonly used ANN architectures for accident prediction include feedforward networks (single-layer and multilayer perceptrons), recurrent neural networks (RNNs), and convolutional neural networks (CNNs), all demonstrating high accuracy in accident-related predictions[18], [19]. Models using multilayer perceptrons with crash-related parameters report accuracies between 73.5% to 74.6% [20], [21], and clustering techniques such as fuzzy c-means further enhance performance in injury severity prediction[20], [22] [1],[20], [22]. More recent applications highlight ANN's versatility:  highlighted roadway geometry on Turkish highways using ANN; [23] employed an ANN–LSTM hybrid for time-series forecasting; [24] reported R ≈ 0.87 in severity modelling, and a Malaysian RNN model reached 71.8% validation accuracy [8]. The shift toward machine learning approaches addresses limitations of traditional statistical models, which often carry inherent assumptions that may lead to inaccurate results[25]. Improved prediction accuracy supports faster medical response times, enhanced traffic management, and alignment with UN 2030 Sustainable Development Goals for reducing road accident fatalities[26].

### 1.3. Optimisation Algorithms for Neural Network Training

In neural network training, the most commonly used optimisation algorithms are first-order methods such as Stochastic Gradient Descent (SGD) and adaptive variants including RMSProp, Adam, and Adagrad. These algorithms have been extensively applied across tasks in image classification, natural language processing, and time series forecasting [3], [4]. Adaptive methods often demonstrate better convergence and stability. Adam and RMSProp, in particular, perform well in both training and evaluation phases [3]. Comparative studies report that while Adam is typically faster to converge [27], SGD with momentum can achieve higher performance on some datasets. Modified approaches such as Hybrid Norm Adam (HN_Adam) have shown improved convergence and accuracy over standard optimisers [28], and hybrid methods like Adaptive Meta Optimisers (ATMO), which blend Adam and SGD, have achieved superior results in classification tasks than using a single optimiser [29]. However, optimisation remains challenging in sequential learning tasks, particularly under class imbalance.

Recent work in optimisation and feature selection further demonstrates the value of hybrid and meta-heuristic approaches for improving learning stability and parameter efficiency. [30] employed backward elimination with grid-search optimisation to enhance Random Forest-based DDoS classification, showing that systematic hyperparameter selection can markedly improve detection performance. In a related study, [31] introduced a hybrid optimisation strategy combining Binary Particle Swarm Optimisation with the Black Widow Algorithm for feature selection and hyperparameter tuning, achieving substantial gains in air-quality prediction accuracy. These findings highlight the broader relevance of integrated feature-selection and optimisation frameworks in domains where complex, high-dimensional data interact with sensitive hyperparameters, an issue that equally affects deep learning models for accident-severity prediction.

Prior work has used gradient magnitudes and per-sample loss as importance or difficulty signals [32]. The authors [30] used gradient normalisation to balance multitask losses, importance-sampling methods derive tractable per-sample gradient-norm upper bounds for efficient sampling, and MentorNet learns data-driven curricula from loss signals to weight [34]. More recently, automatic curriculum/teacher–student approaches have exploited gradient-norm reward signals to guide curriculum generation [35]. However, while these methods reweight losses or select samples to prioritise.

### 1.3.1 Adam Optimiser

The Adam (Adaptive Moment Estimation) optimiser, introduced in 2014, is a popular stochastic optimisation algorithm that adapts learning rates based on lower-order moment estimates for individual parameters[36]. Mathematically expressed as:

$$m_t = \beta_1 m_{t-1} + (1 - \beta_1) g_t \quad \text{(First moment estimate)} \tag{1}$$

$$v_t = \beta_2 v_{t-1} + (1 - \beta_2) g_t^2 \quad \text{(Second moment estimate)} \tag{2}$$

$$\widehat{m_t} = \frac{m_t}{1 - \beta_1^t} \quad \text{(Bias correction for first moment)} \tag{3}$$

$$\widehat{v_t} = \frac{v_t}{1 - \beta_2^t} \quad \text{(Bias correction for second moment)} \tag{4}$$

$$\theta_t = \theta_{t-1} - \frac{\alpha \widehat{m_t}}{\sqrt{\widehat{v_t}} + \varepsilon} \quad \text{(Parameter update)} \tag{5}$$

Where:

$\theta_t$: The updated model parameter at timestamp t.

$\theta_{\{t-1\}}$: The model parameter at timestamp from the previous timestamp$(\mathrm{t} - 1)$.

$g_t$: The gradient of the loss function with respect to the parameter $\theta_{t \text{ at timestep t}}$.

$m_t$: The first moment estimate at time step t, representing the exponentially weighted average of $the$ past gradients.

$v_t$: The second moment estimate at time step t, representing the exponentially weighted average of squared gradients.

$\beta_1$: The decay rate for the first momentum term , typically set to a value close to 1.

$\beta_2$: The decay rate for the first variance term , typically set to a value close to 1.

$\widehat{m}_t$: The bias corrected first moment estimate timestep t.

$\hat{v}_t$: The bias corrected second moment estimate timestep t.

$\varepsilon$: A small constant added to avoid division by zero typically set to a value like $10^{\{-7\}}$

The Adam optimiser has shown significant benefits when applied to Bi-LSTM models across various applications. In fault diagnosis and localisation of power cables, a Bi-LSTM model with Adam optimiser achieved 99.80% accuracy, outperforming traditional machine learning approaches [37]. In Indonesian-Javanese neural machine translation, researchers implemented the Adam optimiser within an LSTM architecture to improve convergence speed and model stability through dynamic learning rate adjustment, resulting in a high BLEU (Bilingual Evaluation Understudy) score of 0.989957 at epoch 2000 [38].

Also, a research study on Bangla literature authorship detection using a Bi-LSTM model optimised with Adam optimiser at a learning rate of 0.0001 achieved the highest testing accuracy of 92% over RMSprop of 85%[39]. For electricity price forecasting, Adam-optimised LSTM neural networks demonstrated improved prediction accuracy compared to traditional methods [40]. These studies collectively highlight the effectiveness of Adam optimiser in enhancing Bi-LSTM model performance across diverse domains, including fault diagnosis, language translation, authorship attribution, and price forecasting.

Despite its popularity, Adam has notable drawbacks, including poor generalisation in some cases, sensitivity to hyperparameters, and issues with convergence, particularly in deep networks [1]. To address Adam's shortcomings, several variants have been proposed: AMSGrad introduces a correction to Adam by ensuring the learning rate does not increase abruptly, preventing convergence issues [1]. Additionally, AdaBound applies dynamic learning rate bounds to improve generalisation. While it stabilises training, it requires careful tuning [41], [42]. Adam with look-ahead (AWL) algorithm enhances the traditional Adam optimiser by incorporating a look-ahead mechanism, which improves convergence properties and performance in various models, including logistic regression and neural networks [43]. However, it still struggles with either convergence speed or overfitting.

While existing optimisers adjust moment estimates, bound learning rates, or reweight samples, they rarely incorporate batch difficulty as a direct optimisation signal. Drawing from curriculum learning and gradient-normalisation principles, this study introduces the Dynamic Batch-Sensitive Adam (DBS-Adam) optimiser. DBS-Adam computes a batch-level difficulty score by combining a normalised gradient norm with a normalised batch loss, and uses this score to dynamically modulate the global learning rate for each batch. Integrated within a Bi-LSTM architecture for accident injury severity prediction, DBS-Adam directly addresses challenges in sequence modelling and class imbalance by enabling more stable updates, improved pattern recognition, and adaptively responsive learning across heterogeneous batches.

### 1.4. This study makes the following contributions:

- Dynamic Batch-Sensitive Optimisation: Introduces DBS-Adam optimiser that adaptively adjusts learning rates based on batch difficulty, enhancing convergence speed and training stability.
- Application of Bi-Directional LSTM: Utilised Bi-LSTM to effectively model temporal dependencies in accident data, enabling better pattern recognition for injury severity.

- Combined Approach to Class Imbalance: Addressed imbalance by integrating SMOTE-ENN with focal loss, improving sensitivity to underrepresented severity classes.

The remainder of this paper is organised as follows: Section 2 presents the proposed methodology, including the DBS-Adam optimiser formulation, Bi-LSTM model architecture, and dataset preprocessing. Section 3 describes the experimental setup and presents results across four progressive experiments, followed by comprehensive comparisons with alternative architectures and state-of-the-art optimisers. Section 4 discusses the findings and their implications for road safety systems. Section 5 concludes the paper and outlines directions for future research.

## 2. Proposed Methodology

The methodological approach of this study is shown in Figure 1 below, which presents the main phases from data collection to accident severity prediction. The process starts with the use of secondary road traffic accident data, followed by preprocessing steps such as data cleaning, feature selection, and transformation. As reported in previous studies, factors such as vehicle type, driver's age, road type, weather conditions, and road alignment are commonly associated with accident injury severity [44], [45], [46], [47]. Based on this evidence, the study retained features supported by the literature for model training. To deal with class imbalance, SMOTE-ENN was applied, and focal loss was used to improve the model's sensitivity to minority classes during training.

The preprocessed dataset was then used to develop a Bi-LSTM model, enhanced with the proposed Dynamic Batch-Sensitive Adam (DBS-Adam) optimiser. Model performance was systematically evaluated through a series of four experiments, each refining the architecture, imbalance handling techniques, and optimisation strategy to achieve improved convergence stability and predictive accuracy.

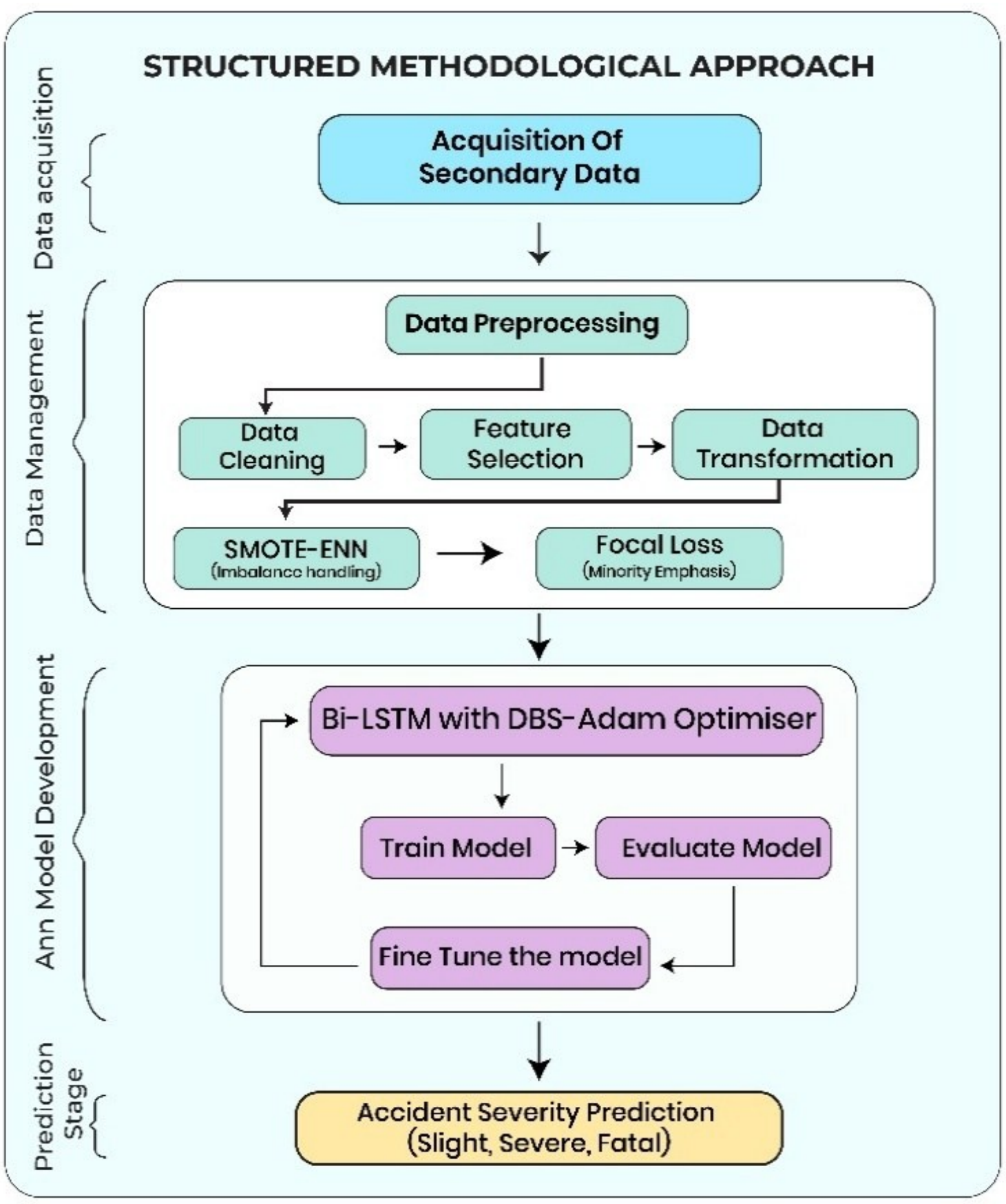

**Figure 1: Design framework for accident severity prediction using Bi-LSTM with the DBS-Adam optimiser.**

In the final stage, the optimised Bi-LSTM framework was applied to classify accident severity into three categories: slight, severe, and fatal injuries. The approach offers a practical way of applying neural network methods to support road safety analysis and emergency response planning.

## 2.1. Proposed Dynamic Batch-Sensitive Adam (DBS-Adam) Optimiser

The proposed Dynamic Batch-Sensitive Adam (DBS-Adam) optimiser introduces an adaptive learning-rate mechanism that responds to batch difficulty during training. Unlike standard Adam, which adjusts learning rates only at the parameter level, DBS-Adam modulates the global learning rate using a difficulty score derived from the normalised gradient norm and batch loss. This approach enables the optimiser to allocate larger updates to challenging batches, which are often enriched with minority-class samples, while reducing updates for easier batches, thereby stabilising convergence and enhancing minority-class learning.

The formulation of DBS-Adam integrates directly with Adam's moment-estimation framework, ensuring computational compatibility whilst introducing a principled method for capturing batch-level complexity.

The gradient norm and batch loss are used in this study as complementary indicators of batch difficulty during training. The gradient norm reflects how strongly the model parameters must change to reduce error for a given batch. Large gradient norms typically arise when the model encounters hard-to-learn patterns, conflicting features, or under-represented classes, indicating that the batch exerts strong optimisation pressure. Prior studies have shown that gradient magnitude is a reliable signal of learning difficulty and optimisation instability, particularly in deep and imbalanced models [32], [33], [35].

The batch loss, in contrast, measures how poorly the model currently predicts the samples within a batch. High loss values indicate that the batch contains misclassified or ambiguous samples, while low loss values suggest that the batch is already well learned. Loss-based difficulty measures have been widely adopted in curriculum learning and hard-example mining to prioritise informative samples and improve minority-class learning [34], [48].

Used together, gradient norm and batch loss provide a more robust estimate of batch difficulty than either signal alone. While batch loss captures predictive error, gradient norm captures optimisation stress. Their combination allows the optimiser to identify batches that are both poorly predicted and difficult to optimise, which are common in imbalanced datasets. This principle has been supported in prior work showing that combining gradient- and loss-based signals leads to more stable and effective learning dynamics [32], [33]. In DBS-Adam, this combined difficulty signal is used to adapt the learning rate at the batch level, enabling larger updates for difficult batches and smaller updates for easier ones.

Although Bi-LSTM combined with Adam generally performs well, its behaviour can be inconsistent on imbalanced sequential datasets. Adam's per-parameter update rule does not account for variations in batch composition, leading to unstable gradient dynamics when minority-class samples occur infrequently or in concentrated batches [4], [36], [49]. These fluctuations weaken performance on rare but critical injury classes [32], [35]. DBS-Adam addresses this limitation by incorporating batch difficulty into the optimisation process, producing more stable updates across heterogeneous batches and enhancing learning in minority-class regimes. Although computationally intensive and costly, our goal is to enable the model to respond more effectively to challenging data and prevent overfitting or stagnation during training. The following subsection details the mathematical formulation and update rule of DBS-Adam.

### 2.1.1 Measuring Batch Difficulty: Gradient Norm and Normalised Loss

In our proposed method, we calculate the difficulty of a mini-batch by combining two complementary batch-level signals: (i) the gradient steepness (Gradient Norm) signal, which captures the local sharpness of the optimisation landscape induced by the batch, and (ii) the batch loss, which measures how poorly the model currently performs on that batch. The equations for deriving our batch difficulty measure are shown in equations [6-12]. The first step is to calculate the mean batch loss, which is expressed as:

$$L_t = \frac{1}{M}\sum_{j=1}^{M} \ell\big(f_\theta(x_j), y_j\big), \tag{6}$$

where; $L_t$ represents the Mean loss over the mini-batch at iteration $t$, $M$ is the Mini-batch size, thus the number of samples in the batch, $(x_j, y_j)$ is used as the input–label pair for sample j in the batch. $\ell(\cdot)$ is the Per-sample loss function and $f_\theta(\cdot)$: Model (parameterised by $\theta$) mapping inputs to outputs. After generating the batch loss value $L_t$, we calculate the full parameter gradient and gradient norm as follows:

$$g_t = \nabla_\theta L_t, G_t = \| g_t \|_2 \tag{7}$$

Where, $g_t$ is the gradient vector of all model parameters with respect to $L_t$, $\theta$ denotes the vector of model parameters, and $G_t$ is the L2 norm of the gradient, which reflects its overall steepness at iteration $t$.

To stabilise fluctuations across batches, we apply exponential moving average (EMA) statistics to normalise both the gradient norm and the batch loss in equation [8]. This produces their z-scores, which represent how far the current values deviate from their recent historical averages.

$$\widetilde{G_t} = \frac{G_t - \mu_G}{\sigma_G + \varepsilon}, \qquad \widetilde{L_t} = \frac{L_t - \mu_L}{\sigma_L + \varepsilon} \tag{8}$$

Here, $\widetilde{G_t}\ and\ \widetilde{L_t}$ denote the z-scored gradient norm and batch loss at iteration $t$. The terms $\mu_G\ and\ \sigma_G$ are the running mean and standard deviation of $G$, while $\mu_L\ and\ \sigma_L$ are the running mean and standard deviation of $L$, both computed using an exponential moving average with a decay factor $\beta$. The constant $\varepsilon$ is included to maintain numerical stability during normalisation.

The normalised values, $\widetilde{G_t}$ $and$ $\widetilde{L_t}$ are then clipped within a fixed range in equation [9] to limit extreme fluctuations and rescaled to the interval [0,1], ensuring that both the gradient norm and batch loss contribute in a controlled and comparable way

$$\hat{G}_t = \frac{clip(\tilde{G}_t, -K, K) + K}{2K}, \qquad \hat{L}_t = \frac{clip(\tilde{L}_t, -K, K) + K}{2K} \tag{9}$$

In this step, $Clip(\cdot, -K, K)$ restricts values to the interval $[-K, K]$, where $K$ is the clipping bound, typically set to 5. The resulting $\hat{G}_t$ $and$ $\hat{L}_t$ are the clipped and rescaled versions of the gradient norm and batch loss, each mapped to the range [0,1]

The clipped and rescaled values $\hat{G}_t$ $and$ $\hat{L}_t$ are then combined into a single linear batch difficulty score, calculated as;

$$D_t = \alpha\, \hat{G}_t + (1 - \alpha)\, \hat{L}_t \tag{10}$$

In this formulation, $D_t$ represents the raw batch difficulty score before any clipping is applied. The parameter $\alpha \in [0,1]$serves as a weighting coefficient that balances the contribution of the normalised gradient norm $\hat{G}_t$ and the normalised batch loss $\hat{L}_t$, with a default value of 0.5.

The batch difficulty score $D_t$ can be understood intuitively as follows: the gradient norm $G_t$ captures how 'steep' the loss landscape is for a given batch. Thus, batches with complex patterns or hard-to-classify samples produce larger gradients as the model struggles to fit them. The batch loss $L_t$ measures the current prediction error. By normalising both metrics through z-scoring and combining them, we obtain a unified difficulty signal. Therefore, batches with both high gradients (indicating complex patterns) and high loss (indicating poor current performance) receive higher difficulty scores, triggering larger learning rate adjustments. This mechanism naturally prioritises learning from challenging minority-class samples, which typically exhibit both characteristics during training.

To prevent instability, the raw difficulty score is clipped to a safe interval, ensuring that the adjusted learning rate remains within reasonable bounds. In this step, $D_t$ is restricted to the range $[D_{min}, D_{max}]$, producing the clipped batch difficulty score.

$$D_t^{\text{clipped}} = clip(D_t, D_{min}, D_{max}) \tag{11}$$

Here, $[D_{min}, D_{max}]$ define the lower and upper bounds, which keep the difficulty scaling factor stable during training. The clipped batch difficulty $D_t^{\text{clipped}}$ score is then used to scale the base learning rate, producing a dynamic learning rate that adapts to the difficulty of each batch in equation [12].

$$\eta_t = \eta_0 \cdot D_t^{\text{clipped}} \tag{12}$$

Here, $\eta_t$ denotes the adjusted learning rate at iteration $t$, computed as the product of the base learning rate $\eta_0$ and the clipped difficulty score $D_t^{\text{clipped}}$. This allows the optimiser to apply larger updates for harder batches and smaller updates for easier ones.

We then use the $D_t^{\text{clipped}}$ to scale the global learning rate $\eta_t = \eta_0 \cdot D_t^{\text{clipped}}$. This linear pipeline produces a single, interpretable difficulty multiplier that balances model steepness and model performance. Once the dynamic learning rate $\eta_t$ is computed, it is applied to update the model weight:

$$\textbf{\textit{Momentum Update}}: \quad m_t = \beta_1 \cdot m_{t-1} + (1 - \beta_1) \cdot g_t \tag{13}$$

$$\textbf{\textit{Second moment Update}}: \quad v_t = \beta_2 \cdot v_{t-1} + (1 - \beta_2) \cdot {g_t}^2 \tag{14}$$

$$\textbf{\textit{Bias Correction momentum}}: \quad \widehat{m}_t = \frac{m_t}{1 - \beta_1^t} \tag{15}$$

$$\textbf{\textit{Bias Correction for second momentum}}: \quad \widehat{v}_t = \frac{v_t}{1 - \beta_2^t} \tag{16}$$

$$\textbf{\textit{Final parameter Update}}: \quad \theta_t = \theta_{t-1} - \eta_t \cdot \frac{\widehat{m}_t}{\sqrt{\widehat{v}_t} + \varepsilon} \tag{17}$$

In this final step, $\theta_t$ denotes the updated model parameters at iteration $t$, and $\eta_t$ is the dynamically adjusted learning rate determined by the batch difficulty. The terms $\widehat{m}_t$ and $\widehat{v}_t$ are the first and second moment estimates of the gradients, and $\varepsilon$ is a small constant added for numerical stability.

**Pseudo-code of the Dynamic Batch-Sensitive Adam (DBS-Adam) optimiser with batch difficulty–based learning rate scaling.**

---

**Algorithm 1: Dynamic Batch-Sensitive Adam (DBS-Adam) Training**

---

**Input:** Dataset D, Model $f(\theta)$, Base learning rate $\eta_0$, Clipping range $[Dmin, Dmax]$, Weight $\alpha$ (balance between gradients and loss), EMA decay $\beta$

**Initialise** $\theta$ (model parameters)
**Initialise** EMA statistics: $\mu G, \sigma G, \mu L, \sigma L$

**for** epoch = 1 to E do
  **for each** batch $B_t$ in **D** do
    # Forward pass
    predictions ← $f\theta(B_t)$
    $L_t$ ← FocalLoss (predictions, labels)

    # Backwards pass (gradients)
    gradients ← $\nabla\theta\ L_t$
    $G_t$ ← mean ($||g||$ for $g$ in gradients)

    # Update EMA statistics
    $\mu G \leftarrow \beta * \mu G + (1-\beta) * G_t$
    $\sigma G \leftarrow \beta * \sigma G + (1-\beta) * |G_t - \mu G|$
    $\mu L \leftarrow \beta * \mu L + (1-\beta) * L_t$
      $\sigma L \leftarrow \beta * \sigma L + (1-\beta) * |L_t - \mu L|$

    # Normalize
    $\hat{G}_t \leftarrow (G_t - \mu G) / (\sigma G + \varepsilon)$
    $\hat{L}_t \leftarrow (L_t - \mu L)/(\sigma L + \varepsilon)$

    # Compute batch difficulty
    $D_t \leftarrow \alpha * \hat{G}_t + (1-\alpha) * \hat{L}_t$
    $D_t \leftarrow clip(D_t, Dmin, Dmax)$

    # Adjust learning rate
    $\eta_t \leftarrow \eta^0 * D_t$

    # Adam update with $\eta_t$
    $\theta \leftarrow AdamUpdate(\theta, gradients, \eta_t)$
  **end for**
**end for**

---

Algorithm 1 shows the pseudo-code of the training steps in the DBS-Adam. Thus, for each batch, the model computes a forward pass and focal loss. Gradients are obtained by backpropagation, and their average norm is measured. Both gradient norm and batch loss are normalised with exponential moving averages to stabilise fluctuations. These values are combined into a batch difficulty score, clipped to a safe range. The base learning rate is then scaled by this score:

easier batches get smaller updates, harder batches larger ones. Adam finally updates the parameters with this adjusted rate. This repeats across batches and epochs, adapting the learning process dynamically.

## 2.2. Positioning of DBS-Adam within Existing Optimisers

First-order optimisation algorithms such as SGD, RMSProp and Adam remain widely used for training neural networks. SGD is simple and effective, but is sensitive to learning-rate choices and often converges slowly on noisy or imbalanced datasets. RMSProp and Adam introduce adaptive step sizes, improving convergence stability across heterogeneous gradients, although Adam may exhibit suboptimal generalisation and erratic behaviour under highly variable gradient magnitudes [1], [3]. Enhanced variants such as AMSGrad, AdamW and AdaBound aim to address these issues by modifying moment estimates or constraining learning-rate dynamics, yet they continue to rely on per-parameter adaptation without considering batch-level difficulty. The proposed DBS-Adam optimiser complements these methods by introducing a dynamic mechanism that scales the global learning rate according to batch difficulty, enabling more consistent optimisation behaviour on imbalanced sequential datasets. This positions DBS-Adam as a batch-sensitive refinement rather than a replacement for existing adaptive optimisers.

## 2.3. The Model Implementation.

We implement a multi-layer Bi-Directional LSTM (Bi-LSTM) network as the predictive backbone [50] . Each node is represented mathematically in equations [18-23], and the pictorial look of the LSTM cell structure is shown in Figure 2:

$$\text{Forget Gate:} f_t = \sigma\left(W_f[h_{t-1}, x_t] + b_f\right) \tag{18}$$

$$\text{Input Gate: } i_t = \sigma(W_i[h_{t-1}, x_t] + b_i) \tag{19}$$

$$\text{Candidate Cell State: } \widetilde{\boldsymbol{C}_t} = \tanh(W_c[h_{t-1}, x_t] + b_c) \tag{20}$$

$$\text{Cell State Update: } (C_t = f_t \cdot C_{t-1} + i_t \cdot \widetilde{C}_t) \tag{21}$$

$$\text{Output Gate: } (o_t = \sigma(W_o[h_{t-1}, x_t] + b_o)) \tag{22}$$

$$\text{Hidden State Update: } (h_t = o_t \cdot \tanh(C_t)) \tag{23}$$

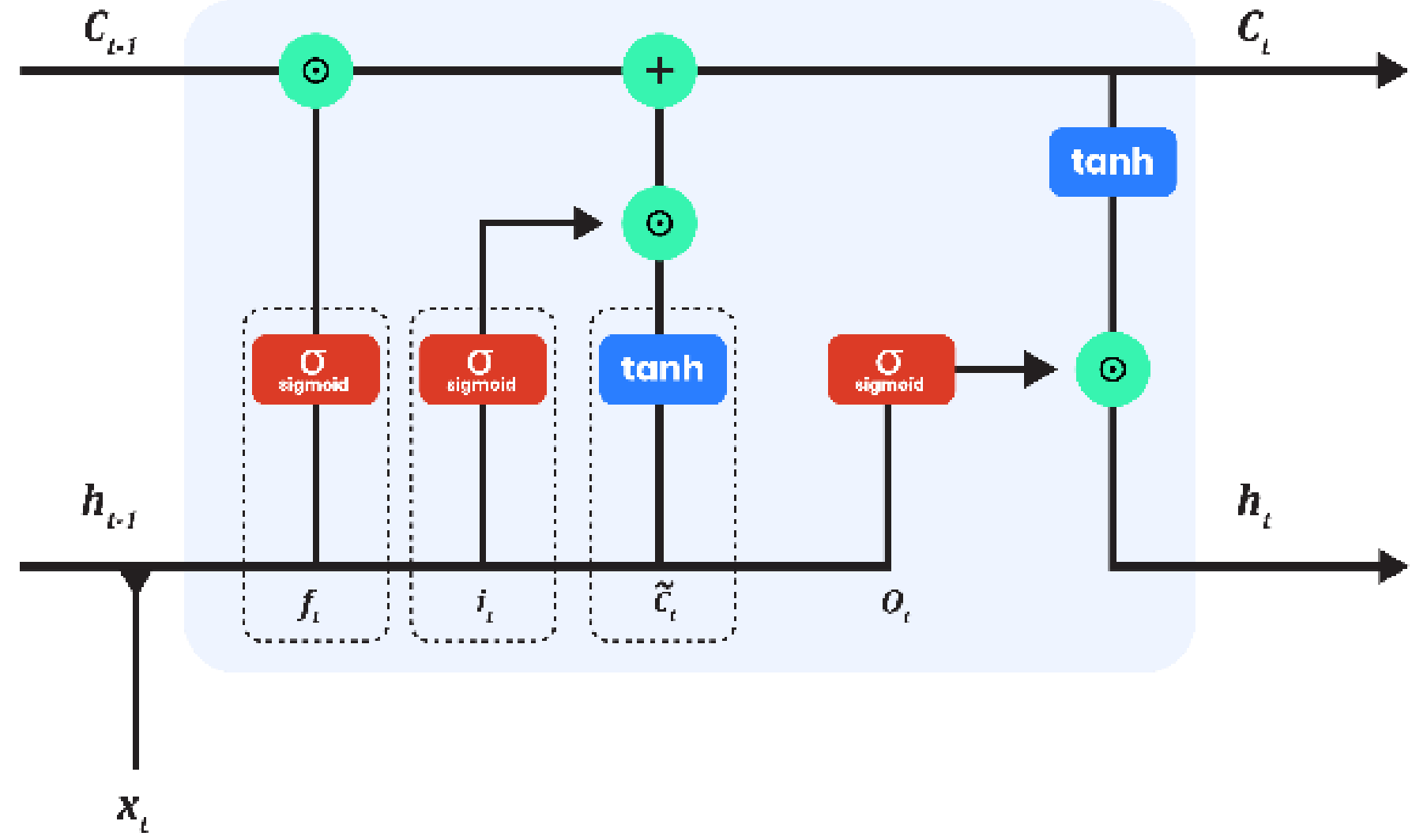

**Figure 2: LSTM Cell Structure**

At time step $t$ *as shown* in Figure 2, the input $x_t$ and the previous hidden state $h_{t-1}$ are combined with the weight matrices $W_f, W_i, W_c, W_o$ and bias terms $b_f, b_i, b_c, b_o$ to compute the forget value $f_t$, input value $i_t$, candidate cell state $\widetilde{C}_t$, and output value $o_t$. The cell state is then updated as $C_t = f_t \cdot C_{t-1} + i_t \cdot \widetilde{C}_t$, where $C_{t-1}$ is the previous cell state. Finally, the hidden state is obtained as $h_t = o_t \cdot \tanh(C_t)$. The functions $\sigma$ (sigmoid) and $tanh$ (hyperbolic tangent) are used for nonlinear transformations, while element-wise multiplication $(\cdot)$ and addition $(+)$ combine the terms.

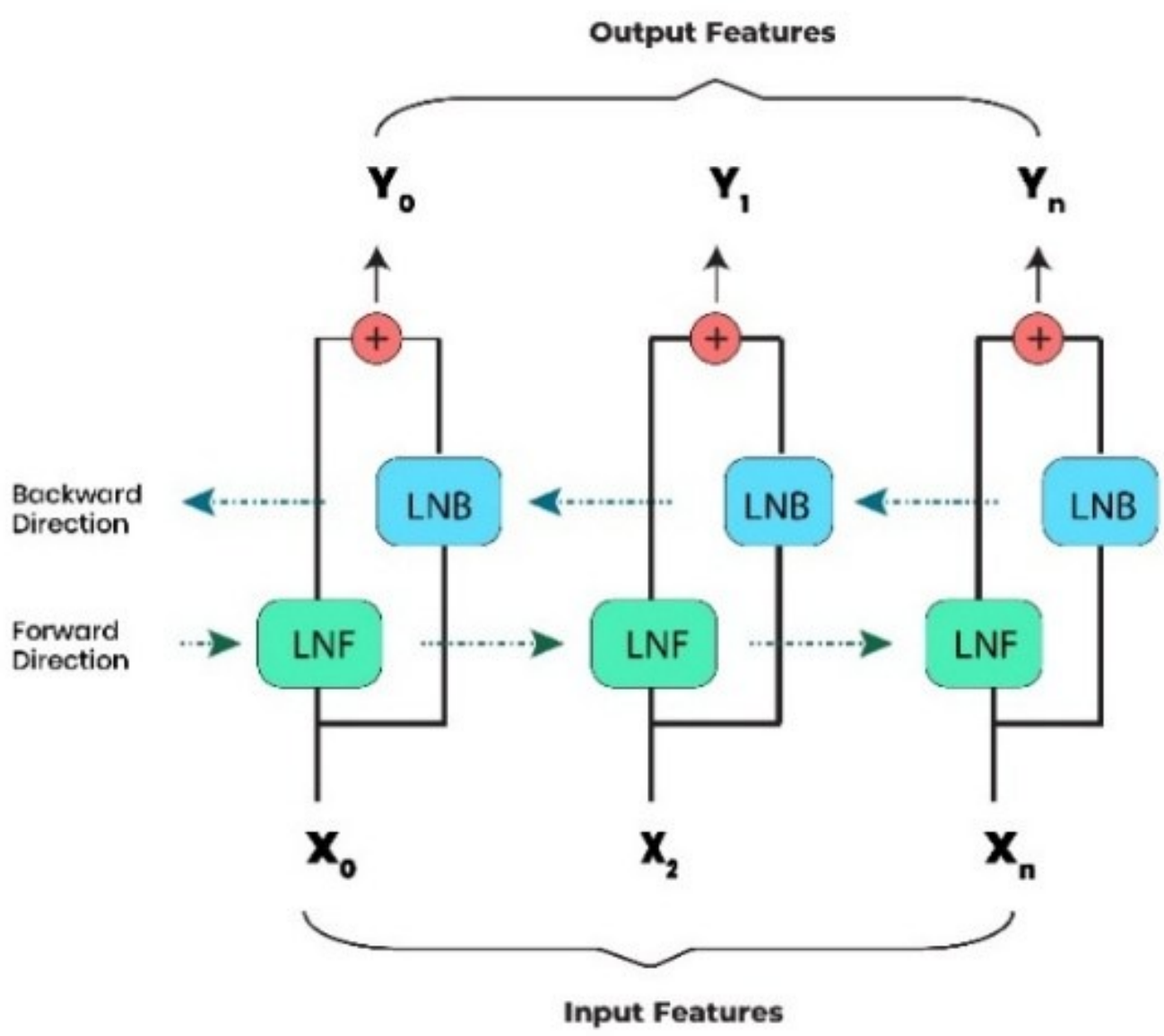

**Figure 3: A Bi-LSTM network processes input in both forward and backwards directions.**

In Figure 3, the Bi-LSTM model accepts a sequence of input features, where each $x_n$ corresponds to a feature vector at a time step $t$. The network processes the sequence in both the forward direction via the $LNF$ and in the backwards direction via $LNB$. For each time step $t$. Then, the forward and backwards hidden states are combined using an additive fusion operation($\oplus$) to yield the final output $Y$. This bidirectional mechanism enables the model to learn dependencies from both past and future contexts, which is crucial for capturing complex patterns in the dataset.

## 2.4. Layer Structure and Parameters

The model comprises several vital layers, each contributing to learning and feature extraction. The layers are structured as follows:

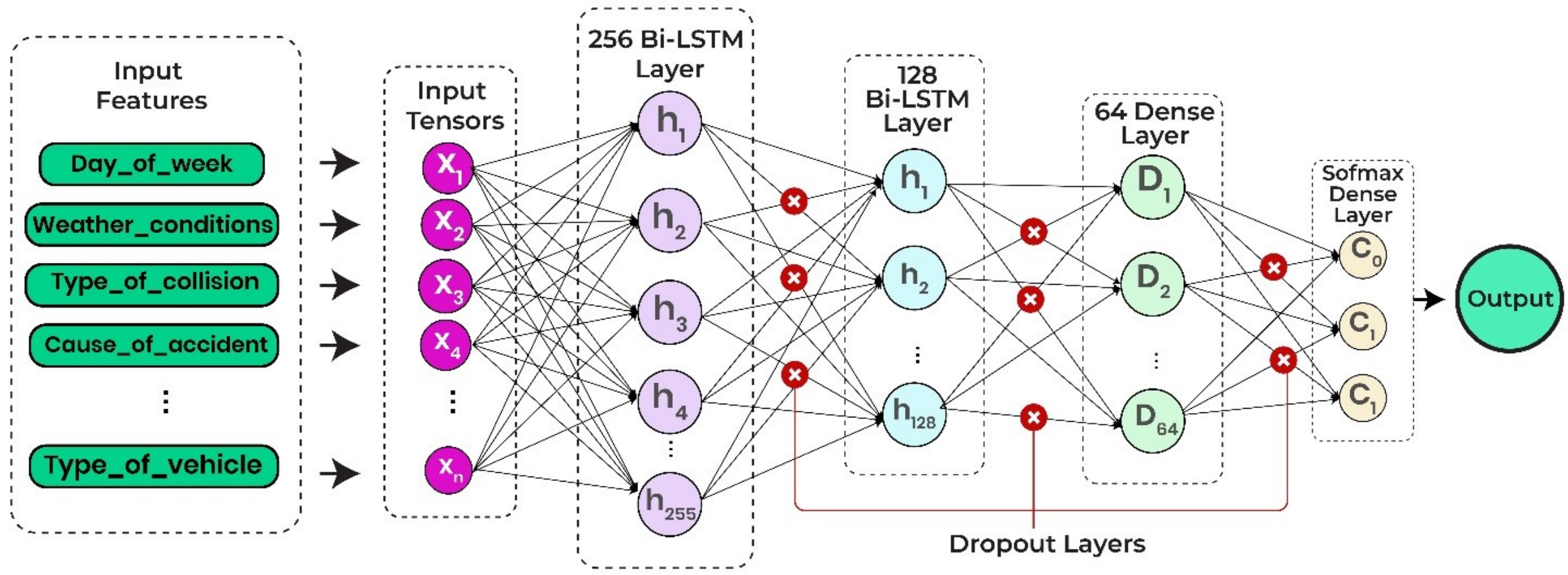

**Figure 4: Proposed Neural network architecture with Bidirectional LSTM layers for accident prediction. The model processes categorical input features through two Bi-LSTM layers (256 and 128 units), followed by a dense layer (64 units) and a softmax output layer with dropout layers.**

In Figure 4, the architecture begins with a Bidirectional LSTM (Bi-LSTM) layer consisting of 256 units, designed to capture both forward and backwards temporal dependencies within sequential accident data. This is followed by a dropout layer, which mitigates overfitting by randomly deactivating a fraction of neurons during training. A second Bi-LSTM layer with reduced dimensionality is stacked to enhance deep feature learning, followed again by a dropout layer for regularisation. These stacked Bi-LSTM layers enable the extraction of important contextual patterns, which are critical for severity classification tasks.

The extracted features are passed through a dense layer with 64 units, followed by dropout, and finally mapped to a 3-class output layer activated by softmax to predict injury severity levels. Dropout regularisation (40%) was applied after each Bi-LSTM and dense layer to prevent overfitting. This rate was selected through experimental evaluation of rates ranging from 30–50%, with 40% providing optimal regularisation whilst maintaining model expressiveness, consistent with established practices in LSTM-based classification where moderate dropout rates (0.2–0.4) are commonly employed [51], [52].

Training this architecture on imbalanced datasets requires careful selection of the loss function to prevent bias toward the majority classes. Two prominent approaches address this challenge: weighted categorical cross-entropy and focal loss. Both loss functions were implemented and evaluated across the experimental configurations described in Section 3. The comparative performance of these approaches informed the final model selection, as detailed in the results section.

### 2.5. Evaluation Metrics

To assess the performance of our Bi-Directional LSTM model with the Dynamic Batch-Sensitive Adam optimiser, we employed a set of well-established evaluation metrics. These include accuracy, precision, recall, and F1-score, each selected to address the class imbalance inherent in the dataset and the critical need to accurately identify all severity classes.

- Accuracy provides an overall correctness measure but can be misleading for imbalanced data.
- Precision and recall were used to evaluate the model's effectiveness in correctly identifying positive instances and its sensitivity across severity classes.
- The F1-score, as a harmonic mean of precision and recall, offered a balanced metric, critical in cases with skewed class distributions.

## 3. Experiment and Results

This section presents the experimental setup and results derived from the bidirectional LSTM-based Artificial Neural Network (ANN) model, enhanced with our novel Dynamic Batch-Sensitive Adam (DBS-Adam) optimiser. The model was developed to predict injury severity from road traffic accidents on major highways, addressing key challenges such as class imbalance and complex temporal dependencies in the dataset.

### 3.1. Dataset and preprocessing

The dataset consists of 12,316 historical records of road traffic accidents. It spans a four-year period (2017–2020) and captures a broad spectrum of accident scenarios in Addis Ababa, Ethiopia [53]. In total, 32 features are included in the original dataset before preprocessing, which encompasses a diverse range of features that capture accident-specific, environmental, vehicular, and temporal factors contributing to injury severity. It can be observed in Table 1 that the dataset exhibits a highly skewed distribution, with a significantly higher number of slight injury cases compared to serious and fatal injuries.

**Table 1 Class Distribution**

| Injury Severity Class | Count | Percentage (%) |
|---|---|---|
| Unknown(unaccounted) (0) | 3956 | 36.124555 |
| Slight Injury (1) | 6294 | 57.474203 |
| Severe Injury (2) | 677 | 6.182084 |
| Fatal Injury (3) | 24 | 0.219158 |

The dataset is imbalanced, with Slight Injury cases accounting for 57.5% of the records, followed by Severe Injuries, while Fatal Injuries are the least represented. In addition, 36.12% of the cases fall under the Unknown category. This class was initially retained because it represents genuine accident reports where injury outcomes were not recorded, and removing it outright would have discarded a large portion of the dataset along with valuable contextual information.

The Unknown category represented records in which injury outcomes were not documented, rather than a clinically meaningful severity level. Retaining this category introduces substantial label noise, as its feature distribution does not correspond to any definable injury outcome. Established work in supervised learning demonstrates that ambiguous or incomplete labels can distort decision boundaries and degrade classifier reliability [54], [55]. The Unknown category represents incomplete documentation rather than a distinct severity level, and including it would force the model to learn data collection artefacts rather than injury severity patterns [54]. This methodological choice aligns with precedent in accident severity research [8], [22], [56] and medical AI systems [57], where predictions must inform actionable decisions such as emergency response and resource allocation. Removing the Unknown category, therefore, does not simplify the task artificially; rather, it restores label validity and allows the model to learn severity-related patterns instead of artefacts of incomplete reporting.

**Data Cleaning and Feature Selection**

The original dataset contained 12,316 accident records. During data cleaning, records with missing or invalid values were removed, reducing the dataset by approximately 1,386 entries. After this stage, the working dataset comprised 10,930 valid records with 32 original features. To prepare the dataset for modelling, one-hot encoding was applied to the categorical variables, which expanded the feature space. As a result, the dimensionality increased from the 32 original features to 63 engineered features. For feature selection, highly correlated variables were identified using a correlation matrix, while Random Forest importance scores and ANOVA F-tests were used to highlight key predictors of accident severity. Important factors included driving experience, vehicle type, time of day, environmental conditions, and location type.

### 3.2. Class Imbalance Handling

Class imbalance in sequential and classification tasks has been addressed through a variety of strategies at the data, loss-function, and model levels. Data-level approaches such as random over-sampling, under-sampling, and hybrid techniques including SMOTE and SMOTE-ENN aim to rebalance class distributions by modifying the training data, though they may introduce redundancy or noise, particularly in temporal datasets [58], [59]. Cost-sensitive and loss-based methods, such as weighted cross-entropy and focal loss, instead assign higher penalties to minority-class errors, improving sensitivity to rare classes but often requiring careful tuning to avoid optimisation instability [48]. At the algorithmic level, curriculum learning and hard-sample mining prioritise difficult examples during training using loss- or gradient-based criteria, enabling more effective learning from minority-class samples [33], [34].

To address the severe class imbalance in the dataset, two data-level resampling techniques were evaluated: Adaptive Synthetic Sampling (ADASYN) and the Synthetic Minority Over-sampling Technique with Edited Nearest Neighbours (SMOTE-ENN). ADASYN adaptively generates synthetic samples for minority classes by focusing on instances that are harder to learn, calculating a density distribution $\Gamma_i$ for each minority sample based on its ratio of majority-class neighbours. The number of synthetic samples $g_i$ generated for the sample $x_i$ is:

$$g_i = \Gamma_i \times G = \frac{r_i}{\sum_{j=1}^{m} r_j} \times G \tag{26}$$

where $G$ is the total synthetic samples to generate and $r_i$ is the majority-neighbour ratio. While this adaptive mechanism aims to strengthen decision boundaries [60].

SMOTE-ENN addresses these limitations through a two-phase approach. First, SMOTE generates synthetic minority samples via interpolation between existing samples and their $k - nearest\ neighbours$:

$$x_{\text{syn}} = x_i + \lambda \times (x_{\text{nn}} - x_i), \quad \lambda \in [0,1] \tag{27}$$

where $x_i$ is a minority sample and $x_{\text{nn}}$ is a randomly selected neighbour. Second, Edited Nearest Neighbours (ENN) removes noisy samples using k-nearest neighbour classification; samples whose predicted class differs from their true label are eliminated. This combined approach balances class distributions while enhancing sample quality [58]. Both techniques were applied to the dataset across different experimental configurations. ADASYN was employed in Experiment 1 with the 4-class dataset, while SMOTE-ENN was applied in subsequent experiments with the refined 3-class dataset, increasing the dataset from 6,995 to 15,403 samples with balanced distribution across severity classes.

On the cost-sensitive and loss-based method, we implemented both focal loss and weighted categorical cross-entropy to handle the imbalance after the data handling techniques. Weighted categorical cross-entropy assigns class-specific penalties inversely proportional to class frequency, formulated as:

$$\mathcal{L}_{\text{weighted}} = -\frac{1}{N}\sum_{i=1}^{N}\sum_{c=1}^{C} w_c \cdot y_{i,c} \cdot \log(\hat{y}_{i,c}) \tag{24}$$

where $w_c$ represents the weight for the class $c$, typically computed as $w_c = \frac{N}{N_c \cdot C}$. This approach directly increases the penalty for minority-class misclassifications, ensuring they contribute more substantially to the overall loss. The primary strength of weighted cross-entropy lies in its simplicity and effectiveness for moderate class imbalance [61].

Focal Loss extends cross-entropy by dynamically down-weighting well-classified examples through a modulating factor, formulated as:

$$\mathcal{L}_{focal} = \alpha(1-\hat{y})^{\gamma} y log(\hat{y}) \tag{25}$$

where $\gamma = 2$ controls the focusing strength and $\alpha = 0.25$ provides class balance. The key strength of focal loss is its instance-level adaptivity—it automatically emphasises hard-to-classify samples regardless of their class, preventing the model from being overwhelmed by easily classified examples [48]. This approach has demonstrated effectiveness in medical diagnosis [62] and traffic accident prediction [63], particularly when class imbalance exceeds $1 : 50$ [64].

Beyond the approaches considered in this study, several alternative techniques have been proposed to address class imbalance in classification and sequential learning tasks. Ensemble-based methods, such as Balanced Random Forests and EasyEnsemble, mitigate imbalance by training multiple classifiers on balanced subsets of the data, thereby improving robustness to minority-class scarcity without altering the base learner [65], [66]. Threshold-moving and decision calibration techniques adjust posterior decision boundaries to favour minority classes, offering a lightweight alternative when retraining is impractical [67]. More recently, representation-level methods, including class-aware feature re-weighting and embedding regularisation, have been used to improve minority-class separability by shaping the learned feature space rather than modifying data or loss functions [68], [69]. Although effective, these methods typically require architectural changes, ensemble construction, or post-hoc adjustments to decisions. In contrast, DBS-Adam focuses on optimisation-level adaptation, enabling improved minority-class learning through batch-aware update dynamics while preserving the original model structure and decision framework.

### 3.3. Target and Feature Encoding and Normalisation

To prepare the dataset for the Bi-LSTM model, one-hot encoding was applied to categorical variables, and numerical features were standardised using z-score normalisation to maintain a zero mean and unit variance, preventing any feature from dominating the learning process. We then split the data into training (80%) and testing (20%) sets using stratified sampling to preserve the class distribution. Finally, target arrays were converted into NumPy arrays with float32 data type, optimising memory usage and computational efficiency for model training.

### 3.4. Experiment Results

To evaluate the Bi-LSTM model's effectiveness in predicting accident severity, we conducted four experiments, progressively refining the approach through resampling techniques, optimisation strategies, and dataset refinements. These experiments aimed to address class imbalance, improve model generalisation, and enhance classification accuracy across all severity levels.

#### 3.4.1 Experiment 1: Bi-LSTM with ADASYN and Focal Loss

This first experiment employed a Bi-LSTM model trained with ADASYN for handling class imbalance and focal loss to emphasise difficult-to-classify samples. ADASYN aimed to generate synthetic instances for minority classes; however, it inadvertently introduced noise and simplistic patterns, impairing the model’s ability to generalise effectively.

The 4-class dataset exhibited severe class imbalance (57.5% Slight, 36.1% Unknown, 6.2% Severe, 0.2% Fatal) as stated earlier. ADASYN was applied to generate synthetic samples for minority classes, increasing the dataset from 10,951 to 17,215 samples. After resampling, the class distribution became more balanced (Table 2), with Fatal Injury increasing from 0.2% to 36.5% and Slight Injury decreasing from 57.5% to 36.6%.

**Table 2: Class Distribution After ADASYN**

| Class | Training Samples | Test Samples | Total | Percentage |
|---|---|---|---|---|
| Unknown | 3165 | 791 | 3956 | 23.0% |
| Slight Injury | 5,035 | 1,259 | 6,294 | 36.6% |
| Severe Injury | 542 | 135 | 677 | 39.% |
| Fatal Injury | 5030 | 1258 | 6288 | 36.5% |
| Total | 13,772 | 3,443 | 17,215 | 100% |

The Model Architecture parameters are listed in Table 3 below:

**Table 3: Experiment 1 model architecture parameters**

| **Parameters** | **Values** |
|---|---|
| Two Bi-LSTM Layers | 64 and 32 Units |
| Dropout Layers | 30% |
| Focal Loss | $\gamma = 2, \alpha = 0.25$ |
| Optimizer | Adam (LR = 0.001) |
| Batch Size | 32 |
| Epochs | 30 (early stopping, patience=6) |

Despite these configurations in Table 3 above, the model achieved only 73.16% test accuracy and a 3.28 test loss, indicating poor generalisation to unseen data.

**Table 4: Classification Report for Experiment 1**

| | Precision | Recall | F1-Score |
|---|---|---|---|
| Unknown | 0.00 | 0.00 | 0.00 |
| Slight Injury | 1.00 | 1.00 | 1.00 |
| Severe Injury | 0.00 | 0.00 | 0.00 |
| Fatal Injury | 0.58 | 1.00 | 0.73 |
| weighted avg. | 0.58 | 0.73 | 0.63 |

The model demonstrated a clear bias toward Slight Injury, which was classified with perfect precision and recall, as shown in Table 4. In contrast, it completely failed to identify Severe Injury and Unknown cases, and only marginally recognised Fatal Injury, indicating the model's inability to capture complex class boundaries. Although focal loss contributed to the focus on harder samples, ADASYN's tendency to introduce noisy synthetic samples ultimately dominated, leading to biased learning. These limitations motivated Experiment 2, which investigated weighted loss as a more effective approach for improving minority class performance and mitigating bias.

### 3.4.2 Experiment 2: Addressing Class Imbalance with SMOTE-ENN and Weighted Loss

In this experiment, SMOTE-ENN (SMOTE + Edited Nearest Neighbours) was used to balance the dataset by oversampling minority classes (Fatal and Severe) while removing noisy samples. Unlike ADASYN used in Experiment 1, this hybrid technique enhances class separability. It minimises misclassified instances through a two-stage process: synthetic sample generation followed by noise filtering using edited nearest neighbours.

The 4-class dataset (including Unknown) was resampled to address the severe class imbalance (57.5% Slight, 36.1% Unknown, 6.2% Severe, 0.2% Fatal). After applying SMOTE-ENN, the class distribution became more balanced, as shown in Table 5.

**Table 5: Class Distribution After SMOTE-ENN**

| Class | Training Samples | Test Samples | Total | Percentage |
|---|---|---|---|---|
| Unknown | 1,968 | 492 | 2,460 | 15.8% |
| Slight Injury | 5,031 | 1,258 | 6,289 | 40.3% |
| Severe Injury | 4,935 | 1,234 | 6,169 | 39.5% |
| Fatal Injury | 556 | 139 | 695 | 4.5% |
| Total | 12,490 | 3,123 | 15,613 | 100% |

To further address class imbalance during training, focal loss was replaced with weighted categorical cross-entropy, assigning higher penalty weights to underrepresented classes. The model architecture parameters are summarised in Table 6.

**Table 6: Model Parameters for Experiment 2**

| Model Architecture Parameters | Values |
|---|---|
| Two Bi-LSTM Layers | 128 and 64 Units |
| Dropout Layers | 40% |
| Loss Function | Weighted Categorical Cross-entropy |
| Class Weights | Unknown: 1.59, Slight: 0.62, Severe: 0.63, Fatal: 5.62 |
| Optimizer | Adam (LR = 0.001) |
| Batch Size | 32 |
| Epochs | 30 (early stopping, patience=6) |

Compared to Experiment 1 (73.16% accuracy), the model demonstrated improved performance with test accuracy increasing to 77.49% with 0.557 test loss and stronger generalisation across most classes, as shown in Table 7.

**Table 7: Classification Report for Experiment 2**

| Class | Precision | Recall | F1-Score |
|---|---|---|---|
| Unknown | 0.43 | 0.51 | 0.47 |
| Slight Injury | 1.00 | 1.00 | 1.00 |
| Severe Injury | 0.93 | 0.64 | 0.76 |
| Fatal Injury | 0.29 | 0.93 | 0.44 |
| Weighted Avg | 0.85 | 0.77 | 0.79 |

Slight Injuries were classified with perfect precision and recall (1.00), demonstrating the model's strong performance on the majority class. Severe Injuries also showed robust performance with high precision (0.93), though recall remained moderate (0.64), indicating some severe cases were misclassified as other categories. However, two critical issues persisted:

Firstly, while recall was high (0.93) on fatal Injury classification, precision remained critically low (0.29), indicating that the model frequently misclassified other severity levels as fatal. This high false-positive rate would result in unnecessary resource mobilisation in operational deployment.

Secondly, the Unknown class exhibited poor precision (0.43) and recall (0.51), with an F1-score of only 0.47. This performance suggests the model struggled to distinguish data collection artefacts from actual severity patterns, as accidents with missing documentation do not share consistent feature patterns with any defined severity level.

The application of SMOTE-ENN and weighted loss contributed to overall improvement over Experiment 1, successfully increasing test accuracy by 4.2 percentage points. However, these techniques did not fully resolve the fundamental challenge posed by the Unknown category's ambiguous nature. The systematic confusion between Fatal and Unknown cases, combined with the Unknown class's inherently inconsistent feature patterns, indicated that including this category compromised the model's ability to learn meaningful severity distinctions.

These findings motivated two key refinements in subsequent experiments: (1) dataset restructuring to exclude the ambiguous Unknown category, enabling the model to focus on clinically defined severity classes, and (2) advanced optimisation strategies to better handle minority class learning and improve convergence stability.

### 3.4.3 Experiment 3: Dataset Refinement and Model Enhancement

As noted in Section 3.1, the original dataset contained an "Unknown" category (36.1%) representing cases where injury outcomes were not documented, a data collection artefact rather than a distinct clinical outcome. Following established principles in supervised classification [54], [55], we excluded this category for the final experiments to ensure the model learned injury severity patterns rather than documentation gaps. This approach aligns with standard practice in accident severity research [8], [22], [56], where predictions must support actionable emergency response decisions.

To empirically validate this decision, we compared the performance of Experiment 2 (4-class with Unknown) against the refined 3-class formulation. After removing Unknown cases, the dataset decreased from 10,930 to 6,995 records. This refined dataset was then balanced using SMOTE-ENN, resulting in 15,403 samples with the distribution shown in Table 8.

**Table 8: Class Distribution After SMOTE-ENN (3-Class Configuration)**

| Class | Training Samples | Test Samples | Total | Percentage |
|---|---|---|---|---|
| Slight Injury | 5,031 | 1,258 | 6,289 | 40.8% |
| Severe Injury | 4993 | 1,249 | 6242 | 40.5% |
| Fatal Injury | 2,298 | 574 | 2,872 | 18.6% |
| Total | 12,322 | 3081 | 15,403 | 100% |

The optimised Bi-LSTM model was trained using the parameters summarised in Table 9.

**Table 9: Model Parameters for Experiment 3**

| Model Architecture Parameters | Values |
|---|---|
| Two Bi-LSTM Layers | 128 and 64 Units |
| Three Dropout Layers | 40% |
| Loss Function | Weighted Categorical Cross-entropy |
| Optimizer | Adam (LR = 0.001) |
| Epochs | 30 (early stopping, patience=6) |

This experiment demonstrated substantial improvement over Experiment 2, achieving a test accuracy of 94.94% and a test loss of 0.1372. Slight and Severe Injuries were classified with near-perfect accuracy, and Fatal Injuries saw significant gains, with an F1-score of 0.88 and a recall of 0.99, as shown in Table 10.

**Table 10 Classification Report for Experiment 3.**

| | precision | recall | f1-score |
|---|---|---|---|
| Slight Injury | 1.00 | 1.00 | 1.00 |
| Severe Injury | 0.99 | 0.88 | 0.94 |
| Fatal Injury | 0.79 | 0.99 | 0.88 |
| weighted avg | 0.96 | 0.95 | 0.95 |

Table 11 compares the performance between Experiment 2 with an unknown class and Experiment 3 without the unknown class:

**Table 11: Performance Comparison between Experiment 2 and Experiment 3**

| Metric | Exp 2 (4-class) | Exp 3 (3-class) | Improvement |
|---|---|---|---|
| Test Accuracy | 77.39% | **94.94%** | +17.55 pp |
| Test Loss | 0.5578 | **0.1372** | –0.4206 (reduction) |
| Precision (Weighted) | 85% | **96%** | +11 pp |
| Recall (Weighted) | 77% | **95%** | +18 pp |
| F1-Score (Weighted) | 79% | **95%** | +16 pp |

The 17.5 percentage-point accuracy improvement empirically demonstrates that excluding the Unknown category substantially enhances the model's ability to distinguish meaningful severity patterns. This is not an artificial simplification; it eliminates noise that interferes with learning the target concept, consistent with established findings on label quality in supervised learning [54].

The enhancements, particularly dataset restructuring, label refinement, and resampling, improved class distinction and reduced misclassification rates. The model showed strong generalisation across training, validation, and test sets, with all metrics consistently above 95%.

However, the test loss of 0.1372 indicated that optimisation was not yet fully effective, even though classification metrics were strong. This outcome reflects some of the known limitations of Adam when working with imbalanced data, particularly in handling difficult batches. These considerations motivated Experiment 4, in which the proposed Dynamic Batch-Sensitive Adam (DBS-Adam) optimiser was applied to adaptively adjust learning rates in response to batch difficulty, thereby improving convergence and further supporting minority-class classification.

### 3.4.4 Experiment 4: Integration of the Dynamic Batch-Sensitive Adam (DBS-Adam) Optimiser

Building on these findings, Experiment 4 applied the proposed Dynamic Batch-Sensitive Adam (DBS-Adam) optimiser to address the higher loss observed in Experiment 3 and to strengthen minority-class prediction further. In this stage, DBS-Adam was evaluated as an adaptive learning-rate method that adjusts the global learning rate based on batch difficulty, measured through normalised gradient norms and batch loss. This mechanism was intended to improve convergence and support more balanced learning across classes. To mitigate dataset imbalance, SMOTE-ENN was applied to increase the representation of Fatal cases while removing noisy samples. In addition, focal loss was incorporated to emphasise hard-to-classify examples, helping to prevent minority classes from being overshadowed by majority classes. The final Bi-LSTM model and its training parameters are summarised in Table 12.

**Table 12: Experiment 4 model and training parameters**

| Model Architecture Parameters | Values |
| --- | --- |
| Two Bi-LSTM Layers | 256 units (first), 128 units (second) |
| Three Dropout Layers | 40% |
| Dense Layer | 64 units (ReLU-Activation Function) |
| Focal loss | $\gamma = 2, \alpha = 0.25$ |
| DBS-Adam | $(Base\ rate = 0.001)$ |
| Epochs | 30 (early stopping, patience=6) |

The model achieved its best performance at epoch 17, with a training accuracy of 95.7% and test accuracy of 94.9%, alongside a very low test loss (0.0088). This indicates strong generalisation and effective handling of class imbalance.

**Table 13: Classification Report for Experiment 4**

| | precision | recall | f1-score |
| --- | --- | --- | --- |
| Slight Injury | 1.00 | 1.00 | 1.00 |
| Severe Injury | 0.99 | 0.88 | 0.93 |
| Fatal Injury | 0.79 | 0.99 | 0.88 |
| weighted avg | 0.96 | 0.95 | 0.95 |

The classification results shown in Table 13 demonstrate that Slight Injuries were predicted with perfect performance across all metrics (Precision, Recall, F1 = 1.00). For Severe Injuries, the model maintained very high precision (0.99), though recall (0.88) suggests that some severe cases were misclassified as fatal. Conversely, Fatal Injuries achieved very high recall (0.99), meaning the model rarely missed a fatal case, although precision (0.79) indicates some over-prediction of fatal cases. This trade-off is particularly valuable in safety-critical domains, where it is better to over-predict fatal cases than to under-predict them. The high macro F1-score of 0.94 shows strong balanced performance across all classes, while the weighted F1-score of 0.95 reflects excellent overall predictive ability. Compared to earlier experiments, the integration of DBS-Adam contributed to: Improved convergence stability via dynamic learning rate modulation, better handling of minority Fatal cases without compromising majority class accuracy and enhanced generalisation, as reflected by both high training and test accuracies and low losses. Thus, DBS-Adam proves effective for accident severity prediction, combining adaptive optimisation with imbalanced data handling to yield robust and practical results.

## 3.5. Performance Comparison and Validation

To comprehensively evaluate the proposed framework, we conducted two main comparative analyses: first, a comparison with alternative neural network architectures employed in earlier accident severity prediction studies; and second, a rigorous assessment of the DBS-Adam optimiser against established adaptive optimisation algorithms. In addition, a sensitivity analysis was performed to examine the stability of DBS-Adam under variations in its key hyperparameters.

### 3.5.1 Comparison with Alternative Neural Network Architectures

To evaluate the effectiveness of the proposed Bi-LSTM model with SMOTE-ENN, Focal Loss, and the DBS-Adam optimiser, we conducted a comparative analysis against several neural network architectures reported in recent accident severity prediction literature. Each baseline model was trained on the same refined 3-class dataset and evaluated using identical performance metrics: accuracy, precision, recall, F1-score, and test loss. The comparison includes:

- A single-layer LSTM with an SGD optimiser and categorical cross-entropy loss [8].
- An LSTM with SMOTE and the Adam optimiser (learning rate = 0.01) [9]
- A baseline Bi-LSTM with standard Adam Optimiser.
- The proposed Bi-LSTM model incorporating SMOTE-ENN, Focal Loss, and the DBS-Adam optimiser.

Performance results are summarised in Table 14.

**Table 14: Comparative Performance of LSTM and Bi-LSTM Models with Various Optimisation and Data Balancing Techniques.**

| Model Description | Accuracy (%) | Precision (%) | Recall (%) | F1-Score (%) | Loss |
|---|---|---|---|---|---|
| Single-layer LSTM architecture with an SGD optimiser, categorical cross-entropy loss [8] | 89 | 81 | 90 | 85 | 0.3451 |
| an LSTM architecture with SMOTE and Adam optimiser (learning rate = 0.01) [9] | 91 | 92 | 91 | 91 | 0.2376 |
| A baseline Bi-LSTM with Adam (Learning Rate 0.001) | 89 | 81 | 90 | 85 | 0.3534 |
| Bi-LSTM + SMOTE-ENN, + Focal Loss + novel DBS-Adam Optimiser | **95.22 ±0.36** | **96.11 ±0.21** | **95.28 ±0.30** | **95.39 ±0.29** | **0.0088± 0.0007** |

The baseline Bi-LSTM with standard Adam and the single-layer LSTM with SGD[8] achieved similar performance, both reaching approximately 89% accuracy with relatively high test loss values (~0.35). These architectures struggled to effectively classify minority classes, reflecting poor handling of class imbalance and weaker generalisation.

The LSTM model with SMOTE and Adam [9] demonstrated improved performance, achieving 91% accuracy and lower test loss (~0.24). The inclusion of SMOTE for handling class imbalance contributed to more balanced precision and recall, though there remained significant room for improvement in generalisation and minority-class sensitivity.

In contrast, the proposed Bi-LSTM framework with SMOTE-ENN, Focal Loss, and DBS-Adam achieved substantially superior performance: 95.22% test accuracy (reported as mean over 5 runs) with the lowest test loss (0.0086). This represents a 4.2 percentage-point improvement over the LSTM+SMOTE baseline [9] and a 6.2 percentage-point improvement over single-layer architectures [8]. The framework's consistent precision (96.11%) and recall (95.28%) across all severity classes confirm its robustness in real-world deployment scenarios. The model's dynamic learning rate adjustments allow it to better adapt to the complexity of each training batch, especially for minority classes such as Fatal Injuries.

### 3.5.2 Comparison with State-of-the-Art Optimisation Algorithms

To rigorously validate the performance of the proposed DBS-Adam optimiser, we conducted a comprehensive comparison against three state-of-the-art adaptive optimisation algorithms: AMSGrad, AdamW, and AdaBound. This comparison isolates the contribution of the optimisation strategy by evaluating all methods under identical experimental conditions.

To ensure statistical rigour and account for stochastic variation in neural network training, each optimiser was evaluated five times using different random seeds (42, 123, 456, 789, 1024). All experiments employed identical configurations shown in Table 15:

**Table 15: Model architecture, training configuration, and evaluation settings used for optimiser comparison experiments.**

| Category | Details |
|---|---|
| **Model Architecture** | Bi-LSTM (256/128 units), 40% dropout, 64-unit dense layer |
| **Training Parameters** | Batch size: 32; Learning rate: 0.001; Early stopping patience: 6; Epoch 30 |
| **Loss Function** | Focal loss ($\gamma = 2$, $\alpha = 0.25$) |
| **Data Preprocessing** | SMOTE-ENN applied to the training set only |
| **Dataset** | Refined 3-class structure: Slight, Severe, Fatal |
| **Evaluation** | Stratified 80/20 train–test split consistent across all seeds |

Results were aggregated to compute the mean and standard deviation across the five runs. Paired t-tests were performed to assess statistical significance ($\alpha = 0.05$), with effect sizes measured using Cohen's d.

Table 16 presents the comparative performance across all optimisers, with results reported as mean ± standard deviation over five independent runs.

**Table 16: Comparative performance of the optimisation algorithms evaluated over five independent runs.**

| Optimizer | Accuracy (%) | Precision (%) | Recall (%) | F1-Score (%) | Test Loss |
|---|---|---|---|---|---|
| **DBS-Adam** | **95.22 ± 0.36** | **96.11 ± 0.21** | **95.28 ± 0.30** | **95.39 ± 0.29** | **0.0086 ± 0.0007** |
| AMSGrad | 95.13 ± 0.27 | 96.01 ± 0.24 | 95.22 ± 0.29 | 95.33 ± 0.27 | 0.0086 ± 0.0006 |
| AdamW | 95.12 ± 0.32 | 96.01 ± 0.21 | 95.18 ± 0.22 | 95.29 ± 0.21 | 0.0086 ± 0.0007 |
| AdaBound | 95.04 ± 0.31 | 95.81 ± 0.29 | 95.09 ± 0.27 | 95.20 ± 0.27 | 0.0086 ± 0.0006 |

To assess whether DBS-Adam's performance improvements were statistically significant, we conducted paired t-tests. Table 17 summarises the statistical comparison for key metrics.

**Table 17 Paired t-test results comparing DBS-Adam with benchmark optimisation algorithms across key performance metrics.**

| Comparison | Metric | Mean Difference | t-statistic | p-value | Cohen's d | Significant (p<0.05) |
|---|---|---|---|---|---|---|
| DBS-Adam vs AMSGrad | Accuracy | +0.09 pp | 1.53 | 0.200 | 0.33 | No |
| | Precision | +0.10 pp | 1.23 | 0.288 | 0.54 | No |
| DBS-Adam vs AdamW | Accuracy | +0.10 pp | 1.43 | 0.226 | 0.32 | No |
| | Precision | +0.10 pp | 1.42 | 0.229 | 0.57 | No |
| DBS-Adam vs AdaBound | Accuracy | +0.18 pp | 2.54 | 0.064 | 0.60 | No |
| | **Precision** | **+0.30 pp** | **3.74** | **0.020** | **1.35** | **Yes** |

From Tables 16 and 17, it can be observed that the DBS-Adam achieved the highest mean performance across all metrics, demonstrating consistent superiority over state-of-the-art optimisers. Whilst differences in accuracy were modest (0.09–0.18 percentage points), the comparison against AdaBound revealed a statistically significant improvement in precision ($p = 0.020, \text{Cohen's } d = 1.35$), indicating a large effect size. This precision advantage is particularly critical for accident severity prediction, where false positives in fatal injury classification can lead to inefficient resource allocation in emergency response systems.

The performance trends across the evaluated metrics reveal several notable observations. DBS-Adam achieved a precision of 96.11%, exceeding AdaBound's 95.81% by 0.30 percentage points, a difference that reached statistical significance. This indicates that the batch-difficulty mechanism reduces false positives more effectively, an important consideration in safety-critical classification where prediction reliability is as important as overall accuracy.

Across the five random seeds, DBS-Adam displayed slightly higher variance ($\pm 0.36\%$) than AMSGrad ($\pm 0.27\%$), yet it consistently attained either the highest or second-highest accuracy in each run, reaching a maximum of 95.62% at seed 1024. This pattern demonstrates the robustness of the adaptive mechanism under differing initialisation conditions.

DBS-Adam reached its optimal validation loss after an average of 17.6 epochs, whereas the other optimisers converged within the range of 9.4–11.2 epochs.

The comparative analysis further shows that DBS-Adam remains competitive with established methods such as AMSGrad and AdamW, while delivering measurable improvements over newer hybrid approaches, particularly AdaBound. The effect sizes ($\text{Cohen's d} = 0.32 - 1.35$) indicate small to large practical differences, with precision demonstrating the largest effect.

Taken together, the results confirm that modulating learning rates according to batch difficulty by integrating gradient-norm information with batch-level loss offers tangible benefits for imbalanced sequential data. Although the improvements over AMSGrad and AdamW do not reach statistical significance at the $p < 0.05$ threshold, DBS-Adam's consistent advantage across all metrics, combined with its statistically significant precision gain over AdaBound, establishes it as a strong optimisation strategy for accident-severity classification. This behaviour aligns with its underlying rationale: by allocating greater update capacity to more difficult batches, often enriched with minority-class samples, the optimiser enhances the modelling of rare but critical classes such as Fatal outcomes.

### 3.5.3 Sensitivity Analysis of DBS-Adam Hyperparameters

A sensitivity analysis was conducted to assess the robustness of DBS-Adam with respect to variations in its key hyperparameters. The exponential moving average coefficient (β) was evaluated at 0.8, 0.9, 0.95 and 0.99, while the difficulty-scaling factor (α) was examined at 0.3, 0.5 and 0.7. All configurations were tested using two random seeds. As shown in Table 18, performance variations across settings were modest. Test accuracy ranged from 0.946 to 0.951, with similarly narrow fluctuations observed for precision and recall.

**Table 18: Sensitivity of DBS-Adam to EMA coefficient (β) and scaling factor (α). Results averaged across two seeds**

| Parameter | Value | Accuracy | Precision | Recall |
|---|---|---|---|---|
| β | 0.80 | 0.9490 | 0.9583 | 0.9490 |
| β | 0.90 | 0.9489 | 0.9577 | 0.9489 |
| β | 0.95 | 0.9487 | 0.9588 | 0.9487 |
| β | 0.99 | 0.9484 | 0.9577 | 0.9484 |
| α | 0.30 | 0.9495 | 0.9583 | 0.9495 |
| α | 0.50 | 0.9489 | 0.9577 | 0.9489 |
| α | 0.70 | 0.9474 | 0.9578 | 0.9474 |

Differences attributable to hyperparameter selection were smaller than the stochastic variation across seeds, indicating stable optimiser behaviour. Among the tested configurations, $\beta = 0.95$ and $\alpha = 0.3$ consistently achieved the strongest overall performance.

## 4. Discussion

The study results demonstrate that the proposed Bi-LSTM framework, enhanced with the Dynamic Batch-Sensitive Adam (DBS-Adam) optimiser, effectively classifies road traffic accident injury severity into three categories: slight, severe, and fatal. Across the four experimental configurations, the model consistently outperformed standard LSTM and Bi-LSTM baselines, particularly in handling class imbalance and achieving superior convergence.

The systematic comparison between 4-class and 3-class formulations demonstrated the critical importance of label quality. Removing the Unknown category resulted in a 17.55 percentage-point accuracy improvement from 77.39% to 94.94% and a 93.6% reduction in test loss. This validates that unambiguous labels on smaller datasets outperform larger datasets with ambiguous target information [54], [55].

Architectural comparison in Table 13 showed the proposed framework achieved 95.22% accuracy compared to 91% for LSTM+SMOTE [9]and 89% for single-layer architectures [8], whilst maintaining balanced precision (96.11%) and recall (95.28%) across all severity classes. Rigorous optimiser comparison (Section 3.5.2) revealed DBS-Adam provides measurable benefits for imbalanced sequential data, with a statistically significant precision improvement over AdaBound (0.30 pp, p=0.020, Cohen's d=1.35). This precision advantage is critical for safety applications, where false-positive fatal predictions trigger unnecessary emergency resource mobilisation.

These results align with earlier findings that deep learning approaches, especially recurrent neural networks, are well-suited for modelling sequential traffic accident data [45], [46]. However, the integration of DBS-Adam, focal loss and SMOTE-ENN extends previous work by improving performance on minority classes, which are often overlooked in accident severity studies. This improvement is particularly relevant in real-world applications, where accurate identification of severe and fatal cases is critical for emergency response.

## 5. Conclusion

This study developed a Bi-Directional LSTM framework for vehicular accident injury-severity prediction, enhanced with Focal Loss, SMOTE-ENN and a Dynamic Batch-Sensitive Adam (DBS-Adam) optimiser. The experiments confirmed that label quality strongly affects model reliability. Removing the ambiguous Unknown category improved accuracy from 77.39% to 94.94%, demonstrating that refined labels contribute more to predictive performance than larger datasets with inconsistent reporting. The final model achieved a mean accuracy of 95.22%, with 96.11% precision and 95.28% recall, outperforming baseline LSTM models and maintaining balanced results across all severity classes.

Comparative analysis with AMSGrad, AdamW and AdaBound showed that DBS-Adam provides consistent gains for imbalanced sequential data. It achieved a statistically significant precision improvement over AdaBound (p = 0.020, Cohen's d = 1.35) and modest but steady improvements over AMSGrad and AdamW. Its batch-difficulty modulation supported stable convergence and enhanced minority-class learning, particularly for Fatal injuries.

The study's conclusions should be considered in light of dataset limitations, including restricted coverage and reduced size after label refinement. Although mitigated through stratified sampling and repeated runs, broader datasets would strengthen generalizability. Future research may extend the difficulty formulation, explore multi-task settings and evaluate operational deployment within emergency-response systems.

## Conflicts of Interest

The authors declare that they have no competing interests.

## Acknowledgements

The authors gratefully acknowledge the helpful feedback and encouragement received from colleagues during the development of this study.

## Authors' Contribution

Daniel Asare Kyei and Maame G. Asante-Mensah were responsible for designing the model and computational framework, analysing the data, performing the calculations, leading the implementation, and drafting the manuscript. Abdul Lateef-Yussiff and Alimatu Saadia-Yussiff supported the planning of the study and provided input during manuscript preparation. Charles Roland Haruna and Derry Emmanuel reviewed the manuscript critically and contributed to refining the final version.

## Funding

The authors declare that no funds, grants, or other support were received during the preparation of this manuscript.

## Data Availability

The accident dataset used in this study is publicly available and can be accessed through the relevant open data repository https://data.mendeley.com/datasets/xytv86278f/1. All processed data and supplementary analysis code supporting the findings of this study are available from the corresponding author upon request.

## References

[1] P. T. Tran and L. T. Phong, "On the Convergence Proof of AMSGrad and a New Version," *IEEE Access*, vol. 7, pp. 61706–61716, 2019, doi: 10.1109/ACCESS.2019.2916341.

[2] Y. Pattanaik *et al.*, "Exploring Optimization Dynamics: Hybrid Approaches Combining Adaptive and Traditional Techniques for Deep Learning Models," in *2025 4th International Conference on Distributed Computing and Electrical Circuits and Electronics (ICDCECE)*, IEEE, Apr. 2025, pp. 1–6. doi: 10.1109/ICDCECE65353.2025.11035942.

[3] R. Zaheer and H. Shaziya, "A Study of the Optimization Algorithms in Deep Learning," in *2019 Third International Conference on Inventive Systems and Control (ICISC)*, IEEE, Jan. 2019, pp. 536–539. doi: 10.1109/ICISC44355.2019.9036442.

[4] D. Soydaner, "A Comparison of Optimization Algorithms for Deep Learning," *Intern J Pattern Recognit Artif Intell*, vol. 34, no. 13, p. 2052013, Dec. 2020, doi: 10.1142/S0218001420520138.

[5] C. Uzondu, S. Jamson, and G. Marsden, "Road safety in Nigeria: unravelling the challenges, measures, and strategies for improvement," *Int J Inj Contr Saf Promot*, vol. 29, no. 4, pp. 522–532, Oct. 2022, doi: 10.1080/17457300.2022.2087230.

[6] M. A. Abdullah and M. A. Yasin, "Road traffic accidents in Pakistan: unveiling the emergency service challenge," *Journal of Basic & Clinical Medical Sciences*, vol. 2, pp. 1–3, Jan. 2024, doi: 10.58398/0002.000007.

[7] S. Birfir, A. Elalouf, and T. Rosenbloom, "Building machine-learning models for reducing the severity of bicyclist road traffic injuries," *Transportation Engineering*, vol. 12, p. 100179, Jun. 2023, doi: 10.1016/j.treng.2023.100179.

[8] M. Sameen and B. Pradhan, “Severity Prediction of Traffic Accidents with Recurrent Neural Networks,” *Applied Sciences*, vol. 7, no. 6, p. 476, Jun. 2017, doi: 10.3390/app7060476.

[9] J. Yuan, M. Abdel-Aty, Y. Gong, and Q. Cai, “Real-Time Crash Risk Prediction using Long Short-Term Memory Recurrent Neural Network,” *Transportation Research Record: Journal of the Transportation Research Board*, vol. 2673, no. 4, pp. 314–326, Apr. 2019, doi: 10.1177/0361198119840611.

[10] T. Sayed and F. Rodriguez, “Accident Prediction Models for Urban Unsignalized Intersections in British Columbia,” *Transportation Research Record: Journal of the Transportation Research Board*, vol. 1665, no. 1, pp. 93–99, Jan. 1999, doi: 10.3141/1665-13.

[11] H. Mensouri, A. Azmani, and M. Azmani, “Towards an Accident Severity Prediction System with Logistic Regression,” in *International Conference on Advanced Intelligent Systems for Sustainable Development*, Springer, 2023, pp. 396–410. doi: 10.1007/978-3-031-26384-2_34.

[12] B. Chong Choo, M. Abdul Razak, M. Z. Mohd Tohir, D. R. Awang Biak, and S. Syam, “An Accident Prediction Model Based on ARIMA in Kuala Lumpur, Malaysia, Using Time Series of Actual Accidents and Related Data,” *Pertanika J Sci Technol*, vol. 32, no. 3, pp. 1103–1122, Apr. 2024, doi: 10.47836/pjst.32.3.07.

[13] J. Zhang, Z. Li, Z. Pu, and C. Xu, “Comparing Prediction Performance for Crash Injury Severity Among Various Machine Learning and Statistical Methods,” *IEEE Access*, vol. 6, pp. 60079–60087, 2018, doi: 10.1109/ACCESS.2018.2874979.

[14] A. Jamal *et al.*, “Injury severity prediction of traffic crashes with ensemble machine learning techniques: a comparative study,” *Int J Inj Contr Saf Promot*, vol. 28, no. 4, pp. 408–427, Oct. 2021, doi: 10.1080/17457300.2021.1928233.

[15] F. E. Sapri, N. S. Nordin, S. M. Hasan, W. F. Wan Yaacob, and S. A. Md Nasir, “Decision Tree Model for Non-Fatal Road Accident Injury,” *Int J Adv Sci Eng Inf Technol*, vol. 7, no. 1, p. 63, Feb. 2017, doi: 10.18517/ijaseit.7.1.1110.

[16] H. R. Al-Masaeid and F. J. Khaled, “Performance of Traffic Accidents’ Prediction Models,” *Jordan Journal of Civil Engineering*, vol. 17, no. 1, pp. 34–44, Jan. 2023, doi: 10.14525/JJCE.v17i1.04.

[17] Z. Li, P. Liu, W. Wang, and C. Xu, “Using support vector machine models for crash injury severity analysis,” *Accid Anal Prev*, vol. 45, pp. 478–486, Mar. 2012, doi: 10.1016/j.aap.2011.08.016.

[18] Md. E. Shaik, Md. M. Islam, and Q. S. Hossain, “A review on neural network techniques for the prediction of road traffic accident severity,” *Asian Transport Studies*, vol. 7, p. 100040, 2021, doi: 10.1016/j.eastsj.2021.100040.

[19] G. Pant, R. Bahuguna, S. Pandey, A. Gehlot, S. P. Yadav, and R. K. Pachauri, “Intelligent Automated Interference for the Protection of Road Safety,” in *2023 International Conference on Computational Intelligence, Communication Technology and Networking (CICTN)*, IEEE, Apr. 2023, pp. 87–91. doi: 10.1109/CICTN57981.2023.10141086.

[20] S. Alkheder, M. Taamneh, and S. Taamneh, “Severity Prediction of Traffic Accident Using an Artificial Neural Network,” *J Forecast*, vol. 36, no. 1, pp. 100–108, Jan. 2017, doi: 10.1002/for.2425.

[21] M. A. Abdel-Aty and H. T. Abdelwahab, “Predicting Injury Severity Levels in Traffic Crashes: A Modeling Comparison,” *J Transp Eng*, vol. 130, no. 2, pp. 204–210, Mar. 2004, doi: 10.1061/(ASCE)0733-947X(2004)130:2(204).

[22] K. Assi, S. M. Rahman, U. Mansoor, and N. Ratrout, “Predicting Crash Injury Severity with Machine Learning Algorithm Synergized with Clustering Technique: A Promising Protocol,” *Int J Environ Res Public Health*, vol. 17, no. 15, p. 5497, Jul. 2020, doi: 10.3390/ijerph17155497.

[23] Patrick Louie Jay R. Federizo, Marriel Bondad-Baet, Arhgy L. Batarlo, Paul Andrei Enriquez, Jimuel Edmon V. Landicho, and Juliana Marie B. Pareja, “Utilization of Artificial Neural Networks (Ann) in Predicting Accidents Within Maharlika Highway San Pablo City, Laguna,” *International Journal of Latest Technology in Engineering Management & Applied Science*, vol. 14, no. 6, pp. 447–459, Jul. 2025, doi: 10.51583/IJLTEMAS.2025.140600049.

[24] M. M. Kunt, I. Aghayan, and N. Noii, “Prediction for traffic accident severity: comparing the artificial neural network, genetic algorithm, combined genetic algorithm and pattern search methods,” *Transport*, vol. 26, no. 4, pp. 353–366, Jan. 2012, doi: 10.3846/16484142.2011.635465.

[25] L. Wahab and H. Jiang, “Severity prediction of motorcycle crashes with machine learning methods,” *International Journal of Crashworthiness*, vol. 25, no. 5, pp. 485–492, Sep. 2020, doi: 10.1080/13588265.2019.1616885.

[26] C. Panda, A. K. Mishra, A. K. Dash, and H. Nawab, “Predicting and explaining severity of road accident using artificial intelligence techniques, SHAP and feature analysis,” *International Journal of Crashworthiness*, vol. 28, no. 2, pp. 186–201, Mar. 2023, doi: 10.1080/13588265.2022.2074643.

[27] C. Ji, “A Survey of Neural Network Optimization Algorithms,” in *2024 IEEE 4th International Conference on Data Science and Computer Application (ICDSCA)*, IEEE, Nov. 2024, pp. 1–7. doi: 10.1109/ICDSCA63855.2024.10859435.

[28] M. Reyad, A. M. Sarhan, and M. Arafa, “A modified Adam algorithm for deep neural network optimization,” *Neural Comput Appl*, vol. 35, no. 23, pp. 17095–17112, Aug. 2023, doi: 10.1007/s00521-023-08568-z.

[29] N. Landro, I. Gallo, and R. La Grassa, “Combining Optimization Methods Using an Adaptive Meta Optimizer,” *Algorithms*, vol. 14, no. 6, p. 186, Jun. 2021, doi: 10.3390/a14060186.

[30] M. S. Sawah, H. Elmannai, A. A. El-Bary, Kh. Lotfy, and O. E. Sheta, “Distributed denial of service (DDoS) classification based on random forest model with backward elimination algorithm and grid search algorithm,” *Sci Rep*, vol. 15, no. 1, p. 19063, May 2025, doi: 10.1038/s41598-025-03868-x.

[31] M. S. Sawah, H. Elmannai, A. A. El-Bary, Kh. Lotfy, and O. E. Sheta, “Improving air quality prediction using hybrid BPSO with BWAO for feature selection and hyperparameters optimization,” *Sci Rep*, vol. 15, no. 1, p. 13176, Apr. 2025, doi: 10.1038/s41598-025-95983-y.

[32] Zhao Chen, Vijay Badrinarayanan, Chen-Yu Lee, and Andrew Rabinovich, “GradNorm: Gradient Normalization for Adaptive Loss Balancing in Deep Multitask Networks,” *Proceedings of the 35th International Conference on Machine Learning*, 2018.

[33] Angelos Katharopoulos and Francois Fleuret, “Not All Samples Are Created Equal: Deep Learning with Importance Sampling,” *Proceedings of the 35th International Conference on Machine Learning*, 2018.

[34] Lu Jiang, Zhengyuan Zhou, Thomas Leung, Li-Jia Li, and Li Fei-Fei, “MentorNet: Learning Data-Driven Curriculum for Very Deep Neural Networks on Corrupted Labels,” *Proceedings of the 35th International Conference on Machine Learning*, 2018.

[35] Ryan Campbell and Junsang Yoon, “Automatic Curriculum Learning with Gradient Reward Signals,” Dec. 2023.

[36] A. Barakat and P. Bianchi, “Convergence and Dynamical Behavior of the ADAM Algorithm for Nonconvex Stochastic Optimization,” *SIAM Journal on Optimization*, vol. 31, no. 1, pp. 244–274, Jan. 2021, doi: 10.1137/19M1263443.

[37] L. Song, “Fault Diagnosis and Localization of Power Cables Using Bi-Directional Long Short Term Memory with Adam Optimizer,” in *2024 4th International Conference on Mobile Networks and Wireless Communications (ICMNWC)*, IEEE, Dec. 2024, pp. 01–05. doi: 10.1109/ICMNWC63764.2024.10872012.

[38] F. I. Putri, A. P. Wibawa, and L. H. Collante, “Refining the Performance of Indonesian-Javanese Bilingual Neural Machine Translation Using Adam Optimizer,” *ILKOM Jurnal Ilmiah*, vol. 16, no. 3, pp. 271–282, Dec. 2024, doi: 10.33096/ilkom.v16i3.2467.271-282.

[39] P. K. Mondal, S. S. Khan, M. T. Imrog, M. A. A. Arman, M. M. Islam, and A. U. H. Rupak, “Exploring Authorial Style in Bangla Literature: LSTM and Bi-LSTM -Based Author Detection,” in *2024 15th International Conference on Computing Communication and Networking Technologies (ICCCNT)*, IEEE, Jun. 2024, pp. 1–9. doi: 10.1109/ICCCNT61001.2024.10725023.

[40] Z. Chang, Y. Zhang, and W. Chen, “Effective Adam-Optimized LSTM Neural Network for Electricity Price Forecasting,” in *2018 IEEE 9th International Conference on Software Engineering and Service Science (ICSESS)*, IEEE, Nov. 2018, pp. 245–248. doi: 10.1109/ICSESS.2018.8663710.

[41] K. Chakrabarti and N. Chopra, “Analysis and Synthesis of Adaptive Gradient Algorithms in Machine Learning: The Case of AdaBound and MAdamSSM,” in *2022 IEEE 61st Conference on Decision and Control (CDC)*, IEEE, Dec. 2022, pp. 795–800. doi: 10.1109/CDC51059.2022.9992512.

[42] J. Liu, J. Kong, D. Xu, M. Qi, and Y. Lu, “Convergence analysis of AdaBound with relaxed bound functions for non-convex optimization,” *Neural Networks*, vol. 145, pp. 300–307, Jan. 2022, doi: 10.1016/j.neunet.2021.10.026.

[43] A. Zhu, Y. Meng, and C. Zhang, “An improved Adam Algorithm using look-ahead,” in *Proceedings of the 2017 International Conference on Deep Learning Technologies*, New York, NY, USA: ACM, Jun. 2017, pp. 19–22. doi: 10.1145/3094243.3094249.

[44] J.-T. Wei, H.-H. Wu, and K.-Y. Kou, “Using Feature Selection to Reduce the Complexity in Analyzing the Injury Severity of Traffic Accidents,” in *2011 International Joint Conference on Service Sciences*, IEEE, May 2011, pp. 329–333. doi: 10.1109/IJCSS.2011.73.

[45] S. Zhang, A. Khattak, C. M. Matara, A. Hussain, and A. Farooq, “Hybrid feature selection-based machine learning Classification system for the prediction of injury severity in single and multiple-vehicle accidents,” *PLoS One*, vol. 17, no. 2, p. e0262941, Feb. 2022, doi: 10.1371/journal.pone.0262941.

[46] Q. Wang, S. Gan, W. Chen, Q. Li, and B. Nie, “A data-driven, kinematic feature-based, near real-time algorithm for injury severity prediction of vehicle occupants,” *Accid Anal Prev*, vol. 156, p. 106149, Jun. 2021, doi: 10.1016/j.aap.2021.106149.

[47] A. Adefabi, S. Olisah, C. Obunadike, O. Oyetubo, E. Taiwo, and E. Tella, "Predicting Accident Severity: An Analysis of Factors Affecting Accident Severity Using Random Forest Model," *International Journal on Cybernetics & Informatics*, vol. 12, no. 6, pp. 107–121, Oct. 2023, doi: 10.5121/ijci.2023.120609.
[48] T.-Y. Lin, P. Goyal, R. Girshick, K. He, and P. Dollár, "Focal loss for dense object detection," in *Proceedings of the IEEE international conference on computer vision*, 2017, pp. 2980–2988.
[49] P. T. Tran and L. T. Phong, "On the Convergence Proof of AMSGrad and a New Version," *IEEE Access*, vol. 7, pp. 61706–61716, 2019, doi: 10.1109/ACCESS.2019.2916341.
[50] A. Graves and J. Schmidhuber, "Framewise phoneme classification with bidirectional LSTM and other neural network architectures," *Neural Networks*, vol. 18, no. 5–6, pp. 602–610, Jul. 2005, doi: 10.1016/j.neunet.2005.06.042.
[51] A. Gajbhiye, S. Jaf, N. Al Moubayed, A. S. McGough, and S. Bradley, "An exploration of dropout with rnns for natural language inference," in *International conference on artificial neural networks*, Springer, 2018, pp. 157–167.
[52] N. Srivastava, G. Hinton, A. Krizhevsky, I. Sutskever, and R. Salakhutdinov, "Dropout: a simple way to prevent neural networks from overfitting," *The journal of machine learning research*, vol. 15, no. 1, pp. 1929–1958, 2014.
[53] T. T. Bedane, "Road traffic accident dataset of addis ababa city," *Addis Ababa*, 2020, doi: 10.17632/xytv86278f.1.
[54] T. Shadbahr *et al.*, "The impact of imputation quality on machine learning classifiers for datasets with missing values," *Communications Medicine*, vol. 3, no. 1, p. 139, Oct. 2023, doi: 10.1038/s43856-023-00356-z.
[55] P. J. García-Laencina, J.-L. Sancho-Gómez, and A. R. Figueiras-Vidal, "Pattern classification with missing data: a review," *Neural Comput Appl*, vol. 19, no. 2, pp. 263–282, Mar. 2010, doi: 10.1007/s00521-009-0295-6.
[56] A. Elalouf, S. Birfir, and T. Rosenbloom, "Developing machine-learning-based models to diminish the severity of injuries sustained by pedestrians in road traffic incidents," *Heliyon*, vol. 9, no. 11, p. e21371, Nov. 2023, doi: 10.1016/j.heliyon.2023.e21371.
[57] M. Bernhardt *et al.*, "Active label cleaning for improved dataset quality under resource constraints," *Nat Commun*, vol. 13, no. 1, p. 1161, Mar. 2022, doi: 10.1038/s41467-022-28818-3.
[58] N. V Chawla, K. W. Bowyer, L. O. Hall, and W. P. Kegelmeyer, "SMOTE: synthetic minority over-sampling technique," *Journal of artificial intelligence research*, vol. 16, pp. 321–357, 2002.
[59] G. E. Batista, R. C. Prati, and M. C. Monard, "A study of the behavior of several methods for balancing machine learning training data," *ACM SIGKDD explorations newsletter*, vol. 6, no. 1, pp. 20–29, 2004.
[60] H. He, Y. Bai, E. A. Garcia, and S. Li, "ADASYN: Adaptive synthetic sampling approach for imbalanced learning," in *2008 IEEE international joint conference on neural networks (IEEE world congress on computational intelligence)*, Ieee, 2008, pp. 1322–1328.
[61] J. Singh *et al.*, "Batch-balanced focal loss: a hybrid solution to class imbalance in deep learning," *Journal of Medical Imaging*, vol. 10, no. 5, p. 51809, 2023.
[62] M. Yeung, E. Sala, C.-B. Schönlieb, and L. Rundo, "Unified focal loss: Generalising dice and cross entropy-based losses to handle class imbalanced medical image segmentation," *Computerized Medical Imaging and Graphics*, vol. 95, p. 102026, 2022.
[63] R. Verma and M. M. Agarwal, "Road Accident Severity Prediction using Adaptive Custom Weight Initialization and Enhanced Focal Loss Integration Technique," *IETE J Res*, pp. 1–13, 2025.
[64] M. Salmi, D. Atif, D. Oliva, A. Abraham, and S. Ventura, "Handling imbalanced medical datasets: review of a decade of research," *Artif Intell Rev*, vol. 57, no. 10, p. 273, 2024.
[65] C. Chen, A. Liaw, and L. Breiman, "Using random forest to learn imbalanced data," *University of California, Berkeley*, vol. 110, no. 1–12, p. 24, 2004.
[66] X.-Y. Liu, J. Wu, and Z.-H. Zhou, "Exploratory undersampling for class-imbalance learning," *IEEE Transactions on Systems, Man, and Cybernetics, Part B (Cybernetics)*, vol. 39, no. 2, pp. 539–550, 2008.
[67] C. Elkan, "The foundations of cost-sensitive learning," in *International joint conference on artificial intelligence*, Lawrence Erlbaum Associates Ltd, 2001, pp. 973–978.
[68] Y. Cui, M. Jia, T.-Y. Lin, Y. Song, and S. Belongie, "Class-balanced loss based on effective number of samples," in *Proceedings of the IEEE/CVF conference on computer vision and pattern recognition*, 2019, pp. 9268–9277.
[69] B. Kang *et al.*, "Decoupling representation and classifier for long-tailed recognition," *arXiv preprint arXiv:1910.09217*, 2019.